\documentclass[a4paper, 10pt, conference]{article}      

\usepackage{graphics} 
\usepackage{epsfig} 
\usepackage{mathptmx} 
\usepackage{times} 
\usepackage{amsmath} 
\usepackage{amssymb}  

\usepackage[letterpaper, margin=2cm]{geometry}

\usepackage[squaren,binary,Gray]{SIunits} 

\usepackage[draft]{todonotes}
\presetkeys{todonotes}{fancyline, color=yellow!30}{}

\usepackage[T1]{fontenc}

\usepackage{booktabs}
\usepackage{array,multirow,makecell}
\newcolumntype{R}[1]{>{\raggedleft\arraybackslash }b{#1}}
\newcolumntype{L}[1]{>{\raggedright\arraybackslash }b{#1}}
\newcolumntype{C}[1]{>{\centering\arraybackslash }b{#1}}
\newcolumntype{M}[1]{>{\centering\arraybackslash}m{#1}}

\usepackage{hyperref}

\title{\LARGE \bf
Classification of Point Cloud Scenes with Multiscale Voxel Deep Network
}

\author{Xavier Roynard, Jean-Emmanuel Deschaud and François Goulette\\ 
 \\
 \{xavier.roynard ; jean-emmanuel.deschaud ; francois.goulette\}@mines-paristech.fr
 \\ 
 \\
 Mines ParisTech, PSL Research University, Centre for Robotics}

\date{}

\begin{document}

\maketitle
\thispagestyle{empty}
\pagestyle{empty}

\begin{abstract}

In this article we describe a new convolutional neural network (CNN) to classify 3D point clouds of urban or indoor scenes. Solutions are given to the problems encountered working on scene point clouds, and a network is described that allows for point classification using only the position of points in a multi-scale neighborhood.

On the reduced-8 Semantic3D benchmark \cite{hackel2017semantic}, this network, ranked second, beats the state of the art of point classification methods (those not using a regularization step).

\end{abstract}

 \begin{figure}[h]
 \includegraphics[width=\linewidth]{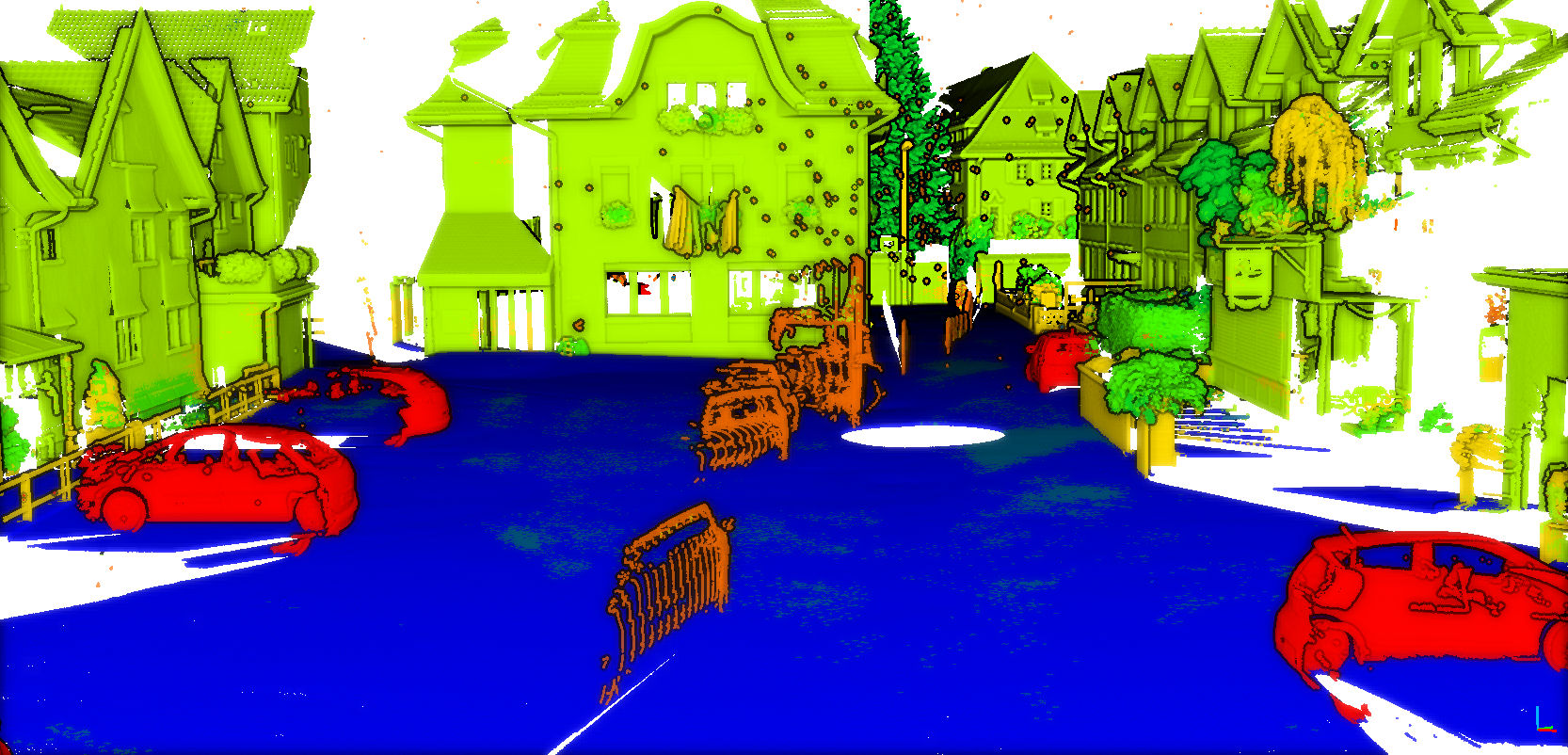}
 \caption{Example of classified point cloud on Semantic3D test set (blue: man-made terrain, cerulean blue: natural terrain, green: high vegetation, light green: low vegetation, chartreuse green: buildings, yellow: hard scape, orange: scanning artefacts, red: cars).\label{fig:semantic}}
 \end{figure}

\section{INTRODUCTION}

Autonomous systems need 3D maps of the world for perception and navigation tasks. These maps can be represented as 3D point clouds with all semantic information needed like lane borders, traffic signs... 

Mobile Laser Scanning (MLS) systems can now scan large areas, like cities or even countries. The produced 3D point clouds can be used as maps for autonomous systems. To do so, the automatic classification of the data is necessary and is still challenging, regards to the number of objects present in an urban scene.


For the object classification task, deep-learning methods work very well on 2D images. The easiest way to transfer these methods to 3D is to use 3D grids. It works well when the data is just one single object \cite{wu20153d}.

But it is much more complicated for the task of point classification of a complete scene (e. g. an urban cloud) made up of many objects of very different sizes and potentially interwoven with each other (e. g. a lamppost passing through vegetation). Moreover, in this kind of scene, there are classes more represented (floor and buildings) than others (pedestrians, traffic signs).

This article proposes both a training method that balances the number of points per class during each epoch, and a 3D CNN capable of effectively learning how to classify scenes containing objects at multiple scales.



\section{STATE OF THE ART}

\subsection{Shallow and Multi-Scale Learning for 3D point cloud classification}

There is a great variety of work for classifying 3D point cloud scenes by shallow learning methods or without learning. Methods can generally be classified into one of the two approaches: classify each point, then group them into objects, or conversely, divide the cloud into objects and classify each object.

 The first approach is followed by \cite{weinmann2015semantic} which classifies each point by calculating simple descriptors such as dimensionality attributes on an optimal neighborhood (minimizing some entropy depending on dimensionality attributes). \cite{hackel2016fast} introduced multi-scale features, computing the same kind of features at different scales to capture both context and local shape around the point. 
 After classifying each point, the points can be grouped into objects by CRF \cite{zhang2015sensor} or by regularization methods \cite{LANDRIEU2017102}.

The segmentation step of the second approach is usually heuristic-based and contains no learning. \cite{aijazi2013segmentation} segments the cloud using super-voxels, \cite{serna2014detection} uses mathematical morphology operators and \cite{roynard2016fast} makes a region growth to extract the soil, then groups the points by related component. After segmentation, objects are classified by computing global descriptors that can be simple geometrical descriptors \cite{serna2014detection}, shape functions \cite{wohlkinger2011ensemble} or histograms of distribution of normals \cite{aldoma2011cad}.

%
%
%
%

\subsection{Deep-Learning for 3D point cloud classification}

Over the past three years, there has been a growing body of work that attempts to adapt deep learning methods or introduces new "deep" approaches to classifying 3D point clouds.

This is well illustrated by the ShapeNet Core55 challenge \cite{yi2017large}, which involved 10 research teams and resulted in the design of new network architectures on both voxel grids and point cloud. The best architectures have beaten the state of the art on the two proposed tasks: part-level segmentation of 3D shapes and 3D reconstruction from single view image.

\subsubsection{on 2D Views of the cloud}\hfill

The most direct approach is to apply 2D networks to images obtained from the point cloud. Among other things, we can think of the following projections:
\begin{itemize}
 \item RGB image rendered from a virtual camera,
 \item depth-map, from a virtual camera,
 \item range image, directly from the sensor,
 \item panorama image\cite{SFIKAS2017},
 \item elevation-map.
\end{itemize}
These methods can be improved by taking multiple views of the same object or scene, and then voting or fusing the results \cite{boulch2017unstructured} (ranked 5th on reduced-8 Semantic benchmark). In addition, these methods greatly benefit from existing 2D expertise and pre-trained networks on image datasets \cite{deng2009imagenet,lin2014microsoft} that contain much more data than point cloud datasets.

\subsubsection{on Voxel Grid}\hfill

The first deep networks used to classify 3D point clouds date from 2015 with VoxNet \cite{maturana2015voxnet}, this network transforms an object instance by filling in an occupancy or density grid and then applies a Convolutional Neural Network (CNN). Later \cite{huang2016point} applied the same type of network to classify clouds of urban points, the network then predicts the class of a point from the occupancy grid of its neighborhood. However, we cannot compare with this architecture because the experimental data has not been published. Best results on ModelNet benchmarks are obtained using deeper CNNs \cite{brock2016generative} based on the architecture of Inception-resnet \cite{szegedy2017inception} and voting on multiple 3D view of objects.


\subsubsection{on Graph}\hfill

Another approach is to use graphs, indeed the raw point cloud having no structure, it is very difficult to derive general information from it. Whereas a graph gives relations of neighborhoods and distances between points and allows for example to make convolutions as in SPGraph \cite{landrieu2017large} or to apply graph-cut methods on CRF as in SEGCloud \cite{tchapmi2017segcloud}.

\subsubsection{on Point Cloud}\hfill

For the time being, there are still quite a few methods that take the point cloud directly as input. These methods have the advantage of working as close as possible to the raw data, so we can imagine that they will be the most efficient in the future.
The first method of this type is PointNet \cite{qi2016pointnet} which gets fairly good results on ModelNet for object instance classification. PointNet is based on the observation that a point cloud is a set and therefore verifies some symmetries (point switching, point addition already in the set...) and is therefore based on the use of operators respecting these symmetries like the global Pooling, but these architectures lose the hierarchical aspect of the calculations that make the strength of the CNN. This gap has been filled with PointNet++ \cite{qi2017pointnet++} which extracts neighborhoods in the cloud, applies PoinNet and groups the points hierarchically to gradually aggregate the information as in a CNN.
Two other approaches are proposed by \cite{engelmann2017exploring} to further account for the context. The first uses PointNet on multiscale neighborhoods, the second uses PointNet on clouds extracted from a 2D grid and uses recurrent networks to share information between grid boxes.

\section{APPROACH}

\subsection{Learning on fully annotated registered point clouds} 

Training on scenes point cloud leads to some difficulties not faced when the point cloud is a single object.

For the point classification task, each point is a sample, so the number of samples per class is very unbalanced (from thousands of points for the class "pedestrian" to tens of millions for the class "ground"). Also with the training method of deep-learning, an Epoch would be to pass through all points of the cloud, which would take a lot of time. Indeed, two very close points have the same neighborhood, and will therefore be classified in the same way. Moreover, for each point one needs to retrieve a neighborhood of the point at a certain scale (or several scales). Even with structures optimized for this task (such as k-d tree or octree) this step can take a lot of time on very large clouds.

We propose a training method that solves these two problems. We randomly select $N$ (for example $1000$) points in each class, then we train on these points mixed randomly between classes, and we renew this mechanism at the beginning of each Epoch.

Once a point $p$ to classify is chosen, we compute a grid of voxels given to the convolutional network by building an occupancy
grid centered on $p$ whose empty voxels contain $0$ and occupied voxels contain $1$. We only use $n\times n\times n$ cubic grids where $n$ is pair, and we only use isotropic space discretization steps $\Delta$. 

\subsection{Data Augmentation and Training}

Some classic data augmentation steps are performed before projecting the 3D point clouds into the voxels grid:
\begin{itemize}
\item{Flip $x$ and $y$ axis, with probability $0.5$}
\item{Random rotation around $z$-axis}
\item{Random scale, between $95\%$ and $105\%$}
\item{Random occlusions (randomly removing points), up to $5\%$}
\item{Random artefacts (randomly inserting points), up to $5\%$}
\item{Random noise in position of points, the noise follows a normal distribution centered in $0$ with standard deviation $0.01\meter$}
\end{itemize}

The cost function used is cross-entropy, and the optimizer used is ADAM \cite{kingma2014adam} with a learning rate of $0.001$ and $\epsilon=10^{-8}$, which are the default settings in most deep-learning libraries.

%
\subsection{Test} \label{subsec:test}

To label a complete point cloud scene, the naive method is to go through all the points of the cloud, and for each point:
\begin{itemize}
 \item look for all the neighboring points that fit into the occupation grid,
 \item create this grid,
 \item infer the class of the point via the pre-trained network. 
\end{itemize}
However, two points very close to each other will have the same neighborhood occupancy grid and therefore the network will predict the same class. A faster test method is therefore to sub-sample the cloud to be tested. This has two beneficial effects: reduce the number of inferences and neighborhood searches, and each neighborhood search takes less time. To infer the point class of the initial cloud, we give each point the class of the nearest point in the subsampled cloud, which can be done efficiently if the subsampling method used retains the correct information. 


\section{NETWORK ARCHITECTURES}

\subsection{3D essential layers}
We denote:
 \begin{itemize}
  \item $Conv(n,k,s,p)$ a convolutional layer that transforms feature maps from previous layer into $n$ new feature maps, with a kernel of size $k\times k\times k$ and stride $s$ and pads $p$ on each side of the grid.
  \item $DeConv(n,k,s,p)$ a transposed convolutional layer that transforms feature maps from previous layer into $n$ new feature maps, with a kernel of size $k\times k\times k$ and stride $s$ and pads $p$ on each side of the grid.
  \item $FC(n)$ a fully-connected layer that transforms the feature maps from previous layer into $n$ feature maps.
  \item $MaxPool(k)$ a layer that aggregates on each feature map every group of 8 neigboring voxels.
  \item $MaxUnPool(k)$ a layer that computes an \textit{inverse} of $MaxPool(k)$.
  \item $ReLU$, $LeakyReLU$ and $PReLU$ common non-linearities used after linear layers as $Conv$ and $FC$. $ReLU(x)$ returns the positive part of $x$, and to avoid null gradient if $x$ is negative, we can add a slight slope which is fixed ($LeakyReLU$) or can be learned ($PReLU$).
  \item $SoftMax$ a non-linearity layer that rescales a tensor in the range $[0,1]$ with sum $1$.
  \item $BatchNorm$ a layer that normalizes samples over a batch.
  \item $DropOut(p)$ a layer that randomly zeroes some of the elements of the input tensor with probability $p$.
 \end{itemize}
 
\subsection{Classification Network Architecture} 
 The choosen network architecture is inspired from \cite{simonyan2014very} that works well in 2D. 
 Our network follows the architecture:
 
\noindent$Conv(32,3,1,0) \rightarrow Conv(32,3,1,0) \rightarrow MaxPool(2) \rightarrow Conv(64,3,1,0) \rightarrow Conv(64,3,1,0) \rightarrow MaxPool(2) \rightarrow FC(1024) \rightarrow FC(N_c)$
 where $N_c$ is the number of classes, and each $Conv$ and $FC$ layer is followed by $BatchNorm \rightarrow PReLU$ and a Squeeze-and-Excitation block \cite{hu2017squeeze} except the last $FC$ layer that is followed by a $SoftMax$ layer. This network takes as input a 3D occupancy grid of size $32\times32\times32$, where each voxel of the grid contains $0$ (empty) or $1$ (occupied) and has a size of $10\centi\meter\times10\centi\meter\times10\centi\meter$.
  
This type of method is very dependent on the space discretization step $\Delta$ selected. Indeed, a small $\Delta$ allows to understand the object finely around the point and its texture (for example to differentiate the natural ground from the ground made by man) but a large $\Delta$ allows to understand the context of the object (for example if it is locally flat and horizontal around the point there can be ambiguity between the ground and the ceiling, but there is no more ambiguity if we add context).
 
Since a 3D scene contains objects at several scales, this type of network can have difficulty classifying certain objects. So we also propose a multiscale version of our network called MS$K$\_DeepVoxScene for the $K$-scales version (or abbreviated in MS$K$\_DVS).
 
\begin{figure}
 \includegraphics[width=\linewidth]{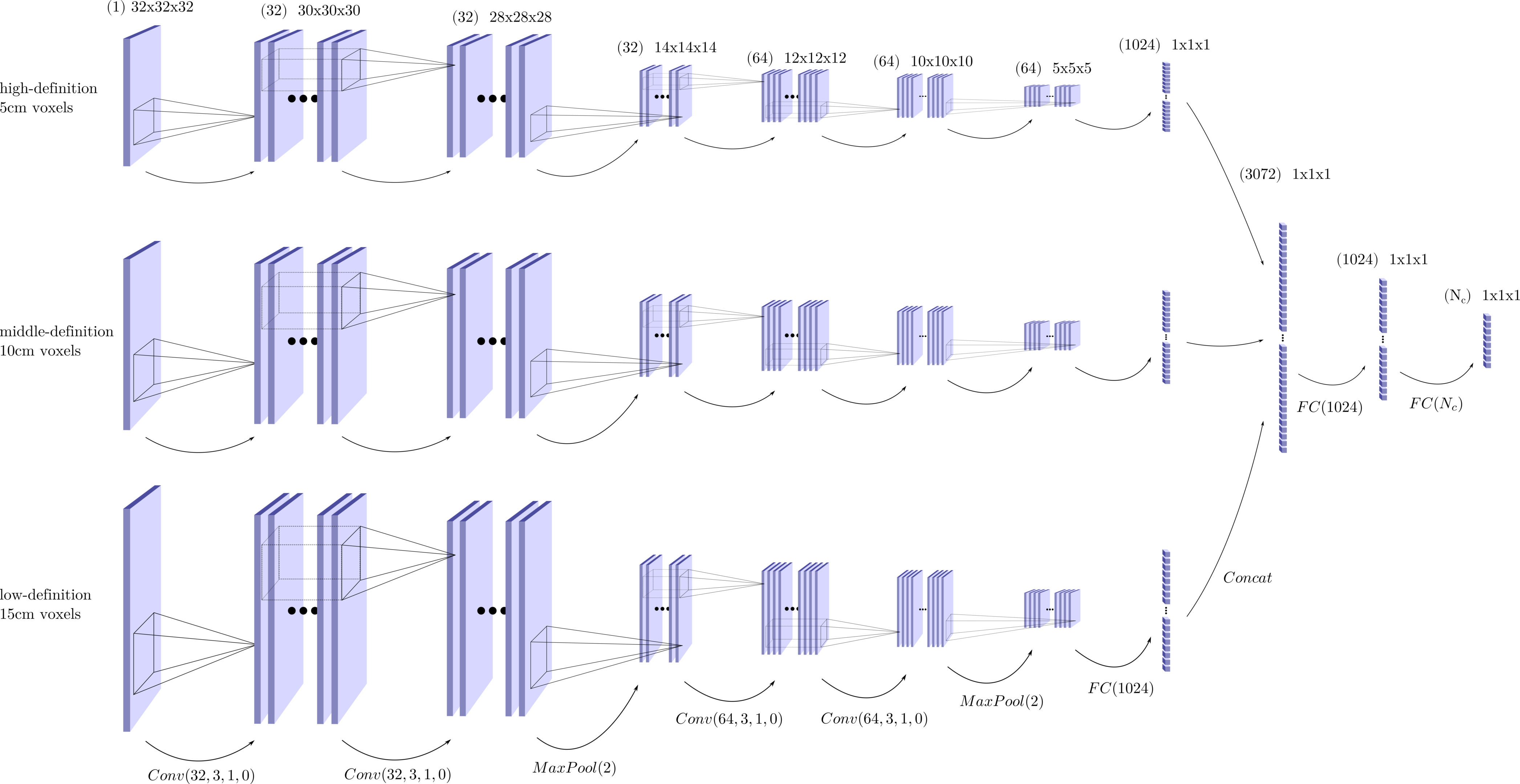}
 \caption{Our Multi-Scale Voxel Network architecture: MS$3$\_DeepVoxScene (all tensors are represented as 2D tensors instead of 3D for simplicity).\label{fig:MultiVoxNetArch}}
\end{figure}

 We take several versions of the previous network without the fully-connected layer. The input of each version is given a grid of the same size $32\times32\times32$, but with different sizes of voxels (for example $5\centi\meter$, $10\centi\meter$ and $15\centi\meter$). We then retrieve a vector of $1024$ characteristics from each version, which we concatenate before giving to a fully-connected classifier layer. See figure \ref{fig:MultiVoxNetArch} for a graphical representation of MS$3$\_DeepVoxScene.
 
\section{EXPERIMENTS}

\subsection{Datasets}

To carry out our experiments we have chosen the 3 datasets of 3D scenes which seem to us the most relevant to train methods of deep-learning, Paris-Lille-3D \cite{roynard2017paris}, S3DIS \cite{armeni_cvpr16} and Semantic3D \cite{hackel2017semantic}. Among the 3D point cloud scenes datasets, these are those with the most area covered and the most variability (see table \ref{tab:datasets}). The covered area is obtained by projecting each cloud on an horizontal plane in pixels of size $10 \centi\meter \times 10 \centi\meter$, then summing the area of all occupied pixels.

\begin{table}\centering
 \begin{tabular}{C{2.8cm}C{2.8cm}C{2.8cm}C{4.0cm}C{2.8cm}}
  \toprule
   \textbf{Name} & \textbf{LiDAR type} & \textbf{Covered Area} & \textbf{Number of points (subsampled)} & \textbf{Number of classes} \\\midrule
   Paris-Lille-3D \cite{roynard2017paris} & multi-fiber MLS  & $55000 \meter^2$ & $143.1\mega$ ($44.0 \mega$) & $9$ \\
   Semantic3D \cite{hackel2017semantic} & static LiDAR & $110000 \meter^2$ & $1660\mega$ ($79.5\mega$) & $8$ \\
   S3DIS \cite{armeni_cvpr16} & MatterPort & $6020 \meter^2$ & $695.9\mega$ ($36.9\mega$) & $13$ \\
  \bottomrule 
 \end{tabular}
 \caption{Comparison of 3D point cloud scenes datasets. Paris-Lille-3D contains $50$ classes but for our experimentations we keep only $9$ coarser classes. In brackets is indicated the number of points after subsampling at $2$ $\centi\meter$. \label{tab:datasets}}
\end{table}

\subsubsection{Paris-Lille-3D}\hfill

\begin{figure}[h]
 \includegraphics[width=0.495\linewidth]{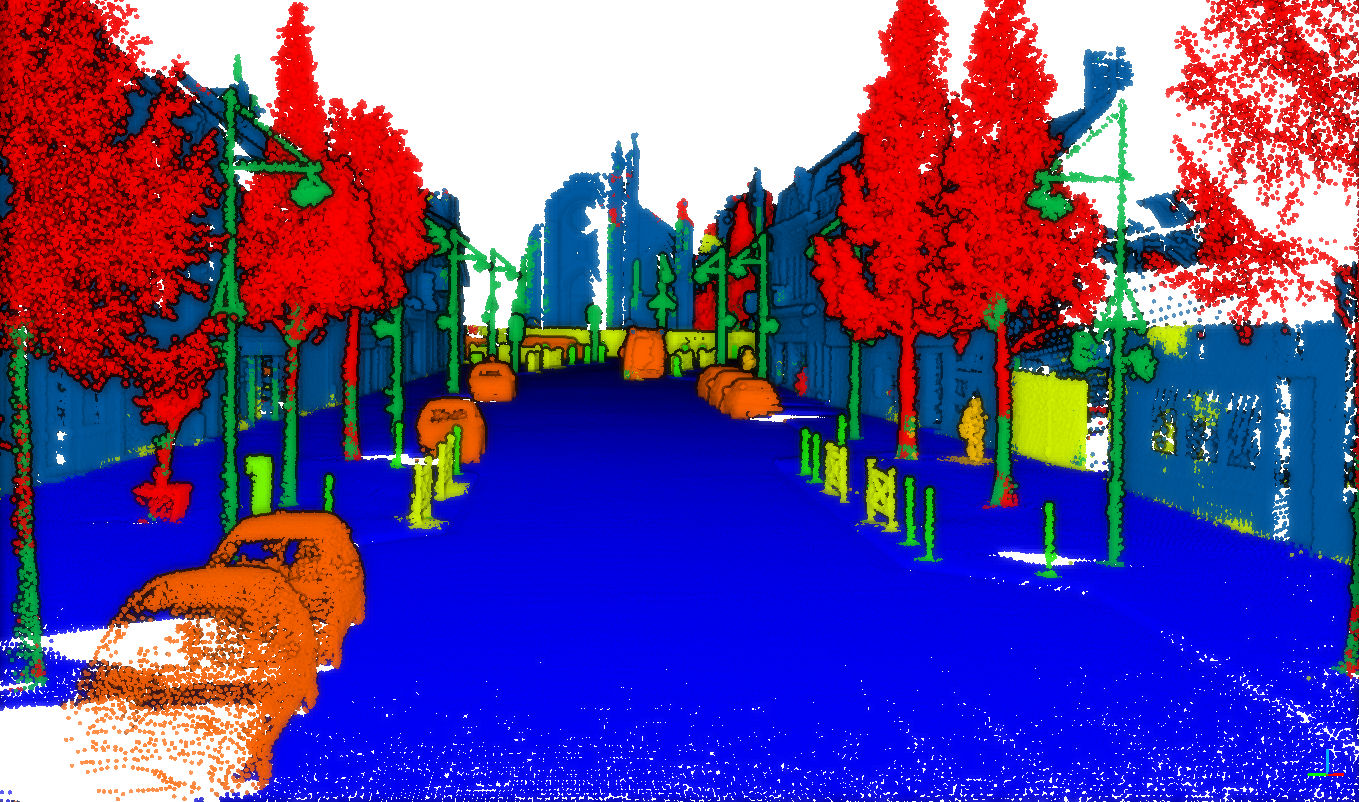} \hfill
 \includegraphics[width=0.495\linewidth]{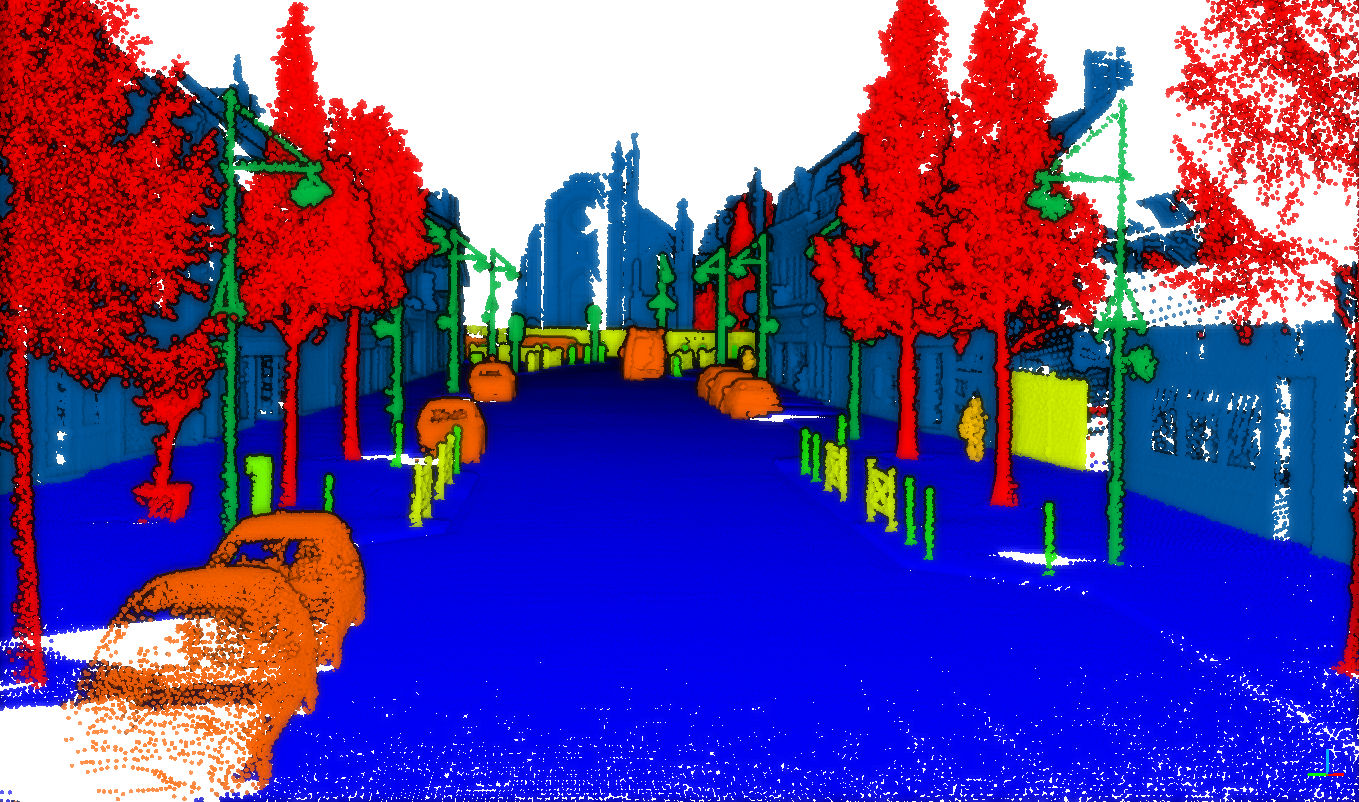}
 \caption{Example of classified point cloud on Paris-Lille-3D dataset. Left: classified with MS$3$\_DVS, right: ground truth (blue: ground, cerulean blue: buildings, dark green: poles, green: bollards, light green: trash cans, yellow: barriers, dark yellow: pedestrians, orange: cars, red: natural).\label{fig:parislille3d}}
\end{figure}

The Paris-Lille-3D dataset consists of $2$ $\kilo\meter$ of 3D point clouds acquired by Mobile Laser Scanning using with a Velodyne HDL-32e mounted on a van. Clouds are georeferenced using IMU and GPS-RTK only, no registration or SLAM methods are used, resulting in a slight noise. Because the scene is scanned at approximately constant speed, the point density is roughly uniform.
The dataset consists of 3 files, one acquired in Paris and two acquired in Lille including \texttt{Lille1.ply} much larger than \texttt{Lille2.ply}. To validate our architectures by $K$-fold method, we cut \texttt{Lille1.ply} into two folds containing the same number of points. 
In addition, this dataset contains 50 classes, some of which only appear in some folds and with very few points. We therefore decide to delete and group together some classes to keep only 9 coarser classes:
\begin{equation}
\begin{matrix}
  \text{ground} & & & \text{buildings} & & & \text{poles} \\
  \text{bollards} & & & \text{trash cans} & & & \text{barriers} \\
  \text{pedestrians} & & & \text{cars} & & & \text{natural}
\end{matrix}\nonumber
\end{equation}
Some qualitative results on Paris-Lille-3D dataset are shown in figure \ref{fig:parislille3d}. 
We can observe that some trunks of trees are classified as poles. It may means that the context is not sufficiently taken into account (even so the $15$ $\centi\meter$ grid is $4.8$ $\meter$ large)
In addition, the ground around objects (except cars) is classified as belonging to the object. One can imagine that cars are not affected by this phenomenon because this class is very present in the dataset.

\subsubsection{Semantic3D}\hfill

The Semantic3D dataset was acquired by static laser scanners, it is therefore more dense than a dataset acquired by MLS as Paris-Lille-3D, but the density of points varies considerably depending on the distance to the sensor. And there are occlusions due to the fact that sensors do not turn around the objects. Even by registering several clouds acquired from different viewpoints, there are still a lot of occlusions. To minimize the problem of very variable density, we subsample the training clouds at $2$ $\centi\meter$. This results in a more uniform density at least close to the sensor and avoids redundant points. After subsampling, the dataset contains $79.5\mega$ points. 
Some qualitative results on Semantic3D dataset are shown in Figure \ref{fig:semantic}.

\subsubsection{S3DIS}\hfill

\begin{figure}[h]
 \includegraphics[width=0.495\linewidth]{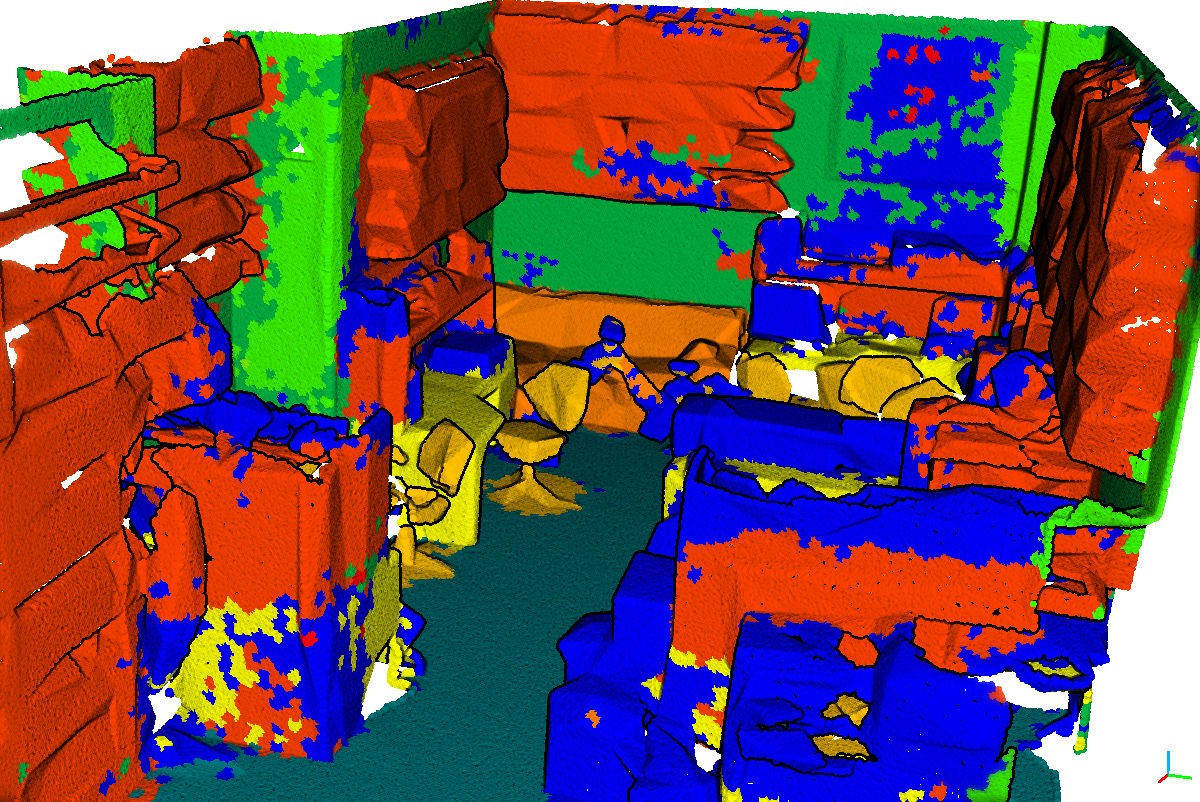}\hfill
 \includegraphics[width=0.495\linewidth]{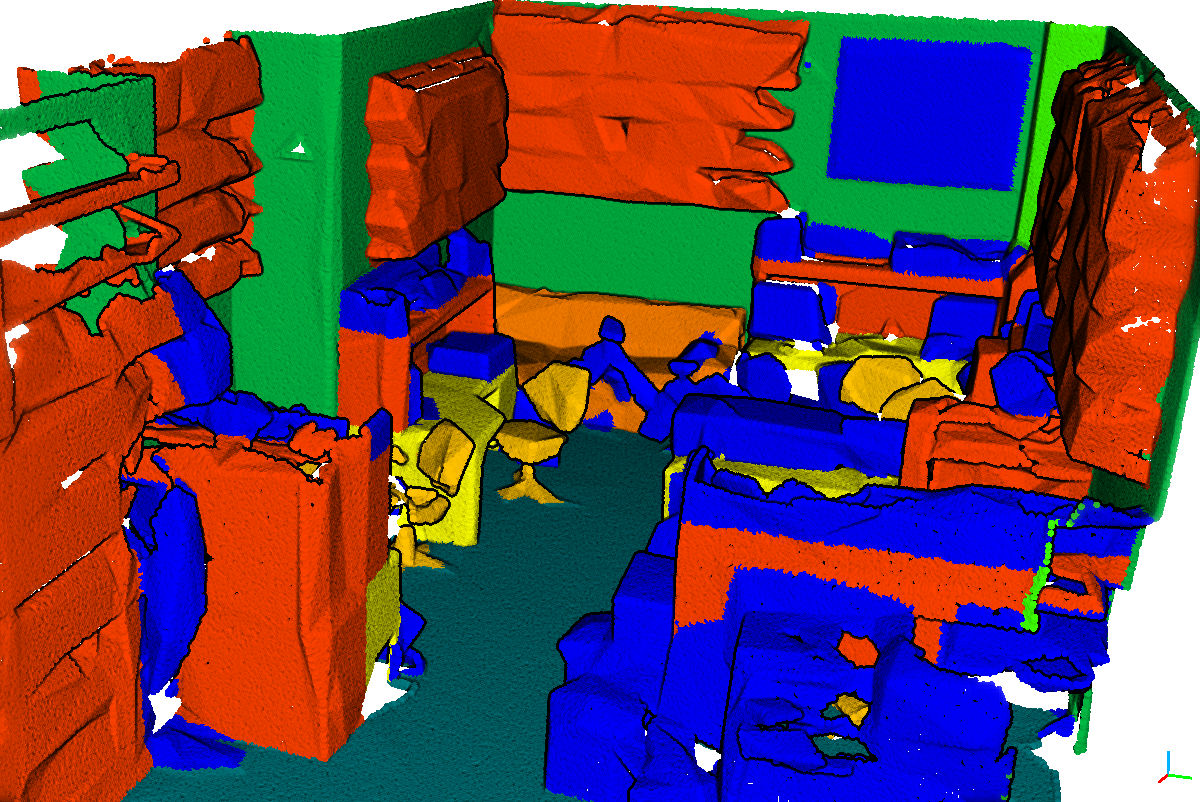}
 \caption{Example of classified point cloud on S3DIS dataset. Left: classified with MS$3$\_DVS, right: ground truth (blue: clutter, cerulean blue: floor, dark green: wall, green: column, dark yellow: chair, light yellow: table, dark orange: bookcase, light orange: sofa).\label{fig:s3dis}}
 \end{figure}

The S3DIS dataset is made up of 6 RGB 3D point cloud scenes taken from 3 different buildings and containing 13 classes. On this dataset we sub-sample the clouds to $2$ $ \centi\meter$ in the same way as for the Semantic3D dataset. After sub-sampling, the entire dataset contains $36,9\mega$ points.
As done in \cite{tchapmi2017segcloud} we compare our results only on fold 5 since this cloud was acquired in another building than the other 5 clouds.
Some qualitative results on S3DIS dataset are shown in figure \ref{fig:s3dis}.
We observe above all that there is a big confusion between the clutter class and the other classes, this can be explained because this class contains objects of all shapes and sizes that can sometimes resemble objects of an existing class. In addition, as on Paris-Lille-3D dataset, the floor around objects such as chairs is classified as belonging to the object instead of the floor. One can guess that during training the network saw only few points on the border between the floor and a chair. 

\subsection{Metrics}
To confirm the interest of multi-scale CNNs, we compare the performance of our two architectures on these three datasets. And on Semantic3D and S3DIS we compare our results with those of the literature.
The metrics used to evaluate performance are the following:
\begin{align} \label{equ:indicators}
 P_c   &= \frac{TP_c}{TP_c + FP_c} \nonumber\\
 R_c   &= \frac{TP_c}{TP_c + FN_c} \nonumber\\
 F1_c  &= \frac{2TP_c}{2TP_c + FP_c + FN_c} = 2\frac{P_c R_c}{P_c + R_c} \nonumber\\
 Acc_c  &= \frac{TP_c}{TP_c + FN_c} \nonumber\\
 IoU_c  &= \frac{TP_c}{TP_c + FP_c + FN_c} \nonumber
\end{align}
Where $P_c$, $R_c$, $F1_c$, $Acc_c$ and $IoU_c$ represent respectively Precision, Recall, F1-score, Accuracy and Intersection-over-Union score of class $c$. And $TP_c$, $TN_c$, $FP_c$ and $FN_c$ are respectively the number of True-Positives, True-Negatives, False-Positives and False-Negatives in class $c$.


\subsection{Comparison with the state of the art}

For a comparison with the state-of-the-art methods on reduced-8 Semantic3D benchmark see table \ref{tab:semantic_test}.
For MS$1$\_DeepVoxScene several resolutions have been tested, and by cross-validation on the Semantic3D training set the $10$ $\centi\meter$ resolution is the one that maximizes validation accuracy. DeepVoxScene's choice of MS$3$\_DeepVoxScene resolution results from this observation, we keep a resolution that obtains good performance in general, and we add a finer resolution of $5$ $\centi\meter$ to better capture the local surface near the point, and a coarser resolution of $15$ $\centi\meter$ to better understand the context of the object to which the point belongs. 
Our method achieves better results than all methods that classify cloud by points (i. e. without regularization). Better results could probably be achieved by adding for exemple a CRF after classification.

\begin{table}
 \sf\centering\small
 \begin{tabular}{cccccccccccc}
 \toprule
  \multirow{10}*{\textbf{Rank}} & \multirow{10}*{\textbf{Method}} & \multirow{9}*{\textbf{Averaged}} & \multirow{9}*{\textbf{Overall}} & \multicolumn{8}{c}{\textbf{Per class IoU}} \\
  \cmidrule(r){5-12} & & \textbf{IoU} & \textbf{Accuracy} & \rotatebox{90}{man-made} \rotatebox{90}{terrain} & \rotatebox{90}{natural} \rotatebox{90}{terrain} & \rotatebox{90}{high} \rotatebox{90}{vegetation} & \rotatebox{90}{low} \rotatebox{90}{vegetation} & \rotatebox{90}{buildings} &  \rotatebox{90}{hard} \rotatebox{90}{scape} & \rotatebox{90}{scanning} \rotatebox{90}{artefacts} & \rotatebox{90}{cars} \\
 \midrule
   1 & SPGraph\cite{landrieu2017large} & $\mathbf{73.2 \%}$ & $\mathbf{94.0 \%}$ & $\mathbf{97.4 \%}$ & $\mathbf{92.6 \%}$ & $\mathbf{87.9 \%}$ & $\mathbf{44.0 \%}$ & $\mathbf{93.2 \%}$ & $31.0 \%$ & $\mathbf{63.5 \%}$ & $76.2 \%$\\
   2 & MS3\_DVS(Ours) & $65.3 \%$ & $88.4 \%$ & $83.0 \%$ & $67.2 \%$ & $83.8 \%$ & $36.7 \%$ & $92.4 \%$ & $\mathbf{31.3 \%}$ & $50.0 \%$ & $\mathbf{78.2 \%}$\\
   3 & RF\_MSSF & $62.7 \%$ & $90.3 \%$ & $87.6 \%$ & $80.3 \%$ & $81.8 \%$ & $36.4 \%$ & $92.2 \%$ & $24.1 \%$ & $42.6 \%$ & $56.6 \%$\\
   4 & SegCloud\cite{tchapmi2017segcloud} & $61.3 \%$ & $88.1 \%$ & $83.9 \%$ & $66.0 \%$ & $86.0 \%$ & $40.5 \%$ & $91.1 \%$ & $30.9 \%$ & $27.5 \%$ & $64.3 \%$\\
   5 & SnapNet\_\cite{boulch2017unstructured} & $59.1 \%$ & $88.6 \%$ & $82.0 \%$ & $77.3 \%$ & $79.7 \%$ & $22.9 \%$ & $91.1 \%$ & $18.4 \%$ & $37.3 \%$ & $64.4 \%$\\\midrule
   9 & MS1\_DVS(Ours) & $57.1 \%$ & $84.8 \%$ & $82.7 \%$ & $53.1 \%$ & $83.8 \%$ & $28.7 \%$ & $89.9 \%$ & $23.6 \%$ & $29.8 \%$ & $65.0 \%$\\
 \bottomrule
 \end{tabular}
  \caption{Top-5 Results on Semantic3D reduced-8 testing set. MS3\_DVS is our MS$3$\_DeepVoxScene with voxel sizes of $5$ $\centi\meter$, $10$ $\centi\meter$ and $15$ $\centi\meter$ and MS1\_DVS is our MS$1$\_DeepVoxScene with voxel size of $10$ $\centi\meter$ (added for comparison with non multi-scale deep network).\label{tab:semantic_test}}
\end{table}


For a comparison with the state-of-the-art methods on S3DIS 5th fold see table \ref{tab:s3dis_test}.
We observe a confusion between the classes wall and board (and more slightly with beam, column, window and door), this is explained mainly because these classes are very similar geometrically and we do not use color. To improve these results, we should not sub-sample the clouds to keep the geometric information thin (such as the table slightly protruding from the wall) and add a $2$ $\centi\meter$ scale in input to the network, but looking for neighbourhoods would then take an unacceptable amount of time.

\begin{table}
 \sf\centering\scriptsize
 \begin{tabular}{ccccccccccccccccc}
 \toprule
  \multirow{10}*{\textbf{Method}} & \multirow{8}*{\textbf{Mean}} & \multirow{8}*{\textbf{Mean}} & \multicolumn{13}{c}{\textbf{Per class IoU (in $\%$)}} \\
  \cmidrule(r){4-16} & \textbf{IoU} & \textbf{Accuracy} & \rotatebox{90}{ceiling} & \rotatebox{90}{floor} & \rotatebox{90}{wall} & \rotatebox{90}{beam} & \rotatebox{90}{column} & \rotatebox{90}{window} & \rotatebox{90}{door} & \rotatebox{90}{chair} & \rotatebox{90}{table} & \rotatebox{90}{bookcase} & \rotatebox{90}{sofa} & \rotatebox{90}{board} & \rotatebox{90}{clutter} \\
 \midrule
  PointNet \cite{qi2016pointnet}          & $41.09 \%$ & $48.98 \%$ & $88.80$ & $ 97.33$ & $ 69.80$ & $\mathbf{0.05}$ & $ 3.92$ & $ 46.26$ & $ 10.76$ & $ 52.61$ & $ 58.93$ & $ 40.28$ & $ 5.85$ & $\mathbf{26.38}$ & $ 33.22$ \\
  MS$3$\_DVS(Ours)                       & $46.32\%$ & $57.93\%$ & $79.03$ & $88.07$ & $53.55$ & $0.00$ & $\mathbf{20.47}$ & $29.01$ & $37.29$ & $68.84$ & $63.72$ & $47.44$ & $\mathbf{61.62}$ & $16.50$ & $36.64$ \\
  SEGCloud \cite{tchapmi2017segcloud}    & $48.92 \%$ & $57.35 \%$ & $90.06$ & $ 96.05$ & $ 69.86$ & $ 0.00$ & $ 18.37$ & $ 38.35$ & $ 23.12$ & $ 75.89$ & $ 70.40$ & $ 58.42$ & $ 40.88$ & $ 12.96$ & $ 41.60$ \\
  SPG \cite{landrieu2017large}           & $\mathbf{54.67 \%}$ & $\mathbf{61.75 \%}$ & $\mathbf{91.49}$ & $\mathbf{97.89}$ & $\mathbf{75.89}$ & $ 0.00$ & $ 14.25$ & $\mathbf{51.34}$ & $\mathbf{52.29}$ & $\mathbf{86.35}$ & $\mathbf{77.40}$ & $\mathbf{65.49}$ & $ 40.38$ & $ 7.23$ & $\mathbf{50.67}$ \\
 \bottomrule
 \end{tabular}
  \caption{Results on S3DIS 5th fold.\label{tab:s3dis_test}}
\end{table}

\subsection{Study of the different architectures}

To evaluate our architecture choices, we tested this classification task by one of the first 3D convolutional networks: VoxNet\cite{maturana2015voxnet}. This allows us both to validate the choices made for the generic architecture of the MS$1$\_DeepVoxScene network and to validate the interest of the multi-scale network. We reimplemented VoxNet using the deep-learning library Pytorch.
See table \ref{tab:compare_methods} for a comparison between VoxNet \cite{maturana2015voxnet}, MS$1$\_DeepVoxScene and MS$3$\_DeepVoxScene on the 3 datasets.

\begin{table}
 \sf\centering
 \begin{tabular}{cccc}
 \toprule
  \textbf{Class} & \textbf{MS$3$\_DVS} & \textbf{MS$1$\_DVS} & \textbf{VoxNet \cite{maturana2015voxnet}} \\
 \midrule
   Paris-Lille-3D & $\mathbf{89.29 \%}$ & $88.23 \%$ & $86.59 \%$ \\
   Semantic3D & $\mathbf{79.36 \%}$ & $74.05 \%$ & $71.66 \%$ \\
   S3DIS & $\mathbf{73.08 \%}$ & $69.36 \%$ & $66.28 \%$ \\
 \bottomrule
 \end{tabular}
  \caption{ Comparison of mean F1 scores of MS$3$\_DVS, MS$1$\_DVS and VoxNet \cite{maturana2015voxnet}. For each dataset, the F1 score is average on all folds.\label{tab:compare_methods}}
\end{table}

See table \ref{tab:parislille_per_class} for a comparison per class between MS$1$\_DeepVoxScene and MS$3$\_DeepVoxScene on Paris-Lille-3D dataset. This shows that the use of multi-scale networks improves the results on some classes, in particular the buildings, barriers and pedestrians classes are greatly improved (especially in Recall), while the car class loses a lot of Precision.

\begin{table}
 \sf\centering
 \begin{tabular}{ccccc}
 \toprule
  \multirow{3}*{\textbf{Class}} & \multicolumn{2}{c}{\textbf{Precision}} & \multicolumn{2}{c}{\textbf{Recall}} \\
  \cmidrule(r){2-3} \cmidrule(r){4-5} & \textbf{MS$3$\_DVS} & \textbf{MS$1$\_DVS} & \textbf{MS$3$\_DVS} & \textbf{MS$1$\_DVS} \\
 \midrule
   ground      & $\mathbf{97.74 \%}$ & $97.08 \%$ & $\mathbf{98.70 \%}$ & $98.28 \%$ \\
   buildings   & $\mathbf{85.50 \%}$ & $84.28 \%$ & $\mathbf{95.27 \%}$ & $90.65 \%$ \\
   poles       & $\mathbf{93.30 \%}$ & $92.27 \%$ & $92.69 \%$ & $\mathbf{94.16 \%}$ \\
   bollards    & $98.60 \%$ & $\mathbf{98.61 \%}$ & $93.93 \%$ & $\mathbf{94.16 \%}$ \\
   trash cans  & $\mathbf{95.31 \%}$ & $93.52 \%$ & $79.60 \%$ & $\mathbf{80.91 \%}$ \\
   barriers    & $\mathbf{85.70 \%}$ & $81.56 \%$ & $\mathbf{77.08 \%}$ & $73.85 \%$ \\
   pedestrians & $\mathbf{98.53 \%}$ & $93.62 \%$ & $\mathbf{95.42 \%}$ & $92.89 \%$ \\
   cars        & $93.51 \%$ & $\mathbf{96.41 \%}$ & $\mathbf{98.38 \%}$ & $97.71 \%$ \\
   natural     & $\mathbf{89.51 \%}$ & $88.23 \%$ & $\mathbf{92.52 \%}$ & $91.53 \%$ \\
 \bottomrule
 \end{tabular}
  \caption{ Per class Precision and Recall averaged on the 4 folds of Paris-Lille-3D dataset. \label{tab:parislille_per_class}}
\end{table}

\section{CONCLUSIONS}

We have proposed both a training method that balances the number of points per class seen during each epoch, as well as a multi-scale CNN that is capable of learning to classify point cloud scenes. This is achieved by both focusing on the local shape of the object around a point and by taking into account the context of the object.

We validated the use of our multi-scale network for 3D scene classification by ranking second on Semantic3D benchmark and by ranking better than state-of-the-art point classification methods (those without regularization).


\addtolength{\textheight}{-12cm}   



%
%

\bibliographystyle{abbrv} 
\bibliography{deepmultiscale_Roynard_2018}

\end{document}